\documentclass[10pt,twocolumn,letterpaper]{article}

\usepackage[pagenumbers]{cvpr} 

\usepackage{graphicx}
\usepackage{amsmath}
\usepackage{amssymb}
\usepackage{booktabs}
\usepackage{multirow}
\usepackage{multicol}
\usepackage{booktabs}
\usepackage{makecell}
\usepackage{bm}
\usepackage[accsupp]{axessibility}

\usepackage[pagebackref,breaklinks,colorlinks]{hyperref}

\usepackage{xcolor}

\usepackage[capitalize]{cleveref}
\crefname{section}{Sec.}{Secs.}
\Crefname{section}{Section}{Sections}
\Crefname{table}{Table}{Tables}
\crefname{table}{Tab.}{Tabs.}

\def\cvprPaperID{****} 
\def\confName{CVPR}
\def\confYear{2023}

\begin{document}

\newcommand{\todo}[1]{{\color{red}$\blacksquare$\textbf{[TODO: #1]}}}

\def\mA{\mathcal{A}}
\def\mB{\mathcal{B}}
\def\mC{\mathcal{C}}
\def\mD{\mathcal{D}}
\def\mE{\mathcal{E}}
\def\mF{\mathcal{F}}
\def\mG{\mathcal{G}}
\def\mH{\mathcal{H}}
\def\mI{\mathcal{I}}
\def\mJ{\mathcal{J}}
\def\mK{\mathcal{K}}
\def\mL{\mathcal{L}}
\def\mM{\mathcal{M}}
\def\mN{\mathcal{N}}
\def\mO{\mathcal{O}}
\def\mP{\mathcal{P}}
\def\mQ{\mathcal{Q}}
\def\mR{\mathcal{R}}
\def\mS{\mathcal{S}}
\def\mT{\mathcal{T}}
\def\mU{\mathcal{U}}
\def\mV{\mathcal{V}}
\def\mW{\mathcal{W}}
\def\mX{\mathcal{X}}
\def\mY{\mathcal{Y}}
\def\mZ{\mathcal{Z}} 

\def\bbN{\mathbb{N}} 
\def\bbR{\mathbb{R}} 
\def\bbP{\mathbb{P}} 
\def\bbQ{\mathbb{Q}} 
\def\bbE{\mathbb{E}}

\def\1n{\mathbf{1}_n}
\def\0{\mathbf{0}}
\def\1{\mathbf{1}}

\def\A{{\bf A}}
\def\B{{\bf B}}
\def\C{{\bf C}}
\def\D{{\bf D}}
\def\E{{\bf E}}
\def\F{{\bf F}}
\def\G{{\bf G}}
\def\H{{\bf H}}
\def\I{{\bf I}}
\def\J{{\bf J}}
\def\K{{\bf K}}
\def\L{{\bf L}}
\def\M{{\bf M}}
\def\N{{\bf N}}
\def\O{{\bf O}}
\def\P{{\bf P}}
\def\Q{{\bf Q}}
\def\R{{\bf R}}
\def\S{{\bf S}}
\def\T{{\bf T}}
\def\U{{\bf U}}
\def\V{{\bf V}}
\def\W{{\bf W}}
\def\X{{\bf X}}
\def\Y{{\bf Y}}
\def\Z{{\bf Z}}

\def\a{{\bf a}}
\def\b{{\bf b}}
\def\c{{\bf c}}
\def\d{{\bf d}}
\def\e{{\bf e}}
\def\f{{\bf f}}
\def\g{{\bf g}}
\def\h{{\bf h}}
\def\i{{\bf i}}
\def\j{{\bf j}}
\def\k{{\bf k}}
\def\l{{\bf l}}
\def\m{{\bf m}}
\def\n{{\bf n}}
\def\o{{\bf o}}
\def\p{{\bf p}}
\def\q{{\bf q}}
\def\r{{\bf r}}
\def\s{{\bf s}}
\def\t{{\bf t}}
\def\u{{\bf u}}
\def\v{{\bf v}}
\def\w{{\bf w}}
\def\x{{\bf x}}
\def\y{{\bf y}}
\def\z{{\bf z}}

\def\balpha{\mbox{\boldmath{$\alpha$}}}
\def\bbeta{\mbox{\boldmath{$\beta$}}}
\def\bdelta{\mbox{\boldmath{$\delta$}}}
\def\bgamma{\mbox{\boldmath{$\gamma$}}}
\def\blambda{\mbox{\boldmath{$\lambda$}}}
\def\bsigma{\mbox{\boldmath{$\sigma$}}}
\def\btheta{\mbox{\boldmath{$\theta$}}}
\def\bomega{\mbox{\boldmath{$\omega$}}}
\def\bxi{\mbox{\boldmath{$\xi$}}}
\def\bnu{\mbox{\boldmath{$\nu$}}}                                  
\def\bphi{\mbox{\boldmath{$\phi$}}}
\def\bmu{\mbox{\boldmath{$\mu$}}}

\def\bDelta{\mbox{\boldmath{$\Delta$}}}
\def\bOmega{\mbox{\boldmath{$\Omega$}}}
\def\bPhi{\mbox{\boldmath{$\Phi$}}}
\def\bLambda{\mbox{\boldmath{$\Lambda$}}}
\def\bSigma{\mbox{\boldmath{$\Sigma$}}}
\def\bGamma{\mbox{\boldmath{$\Gamma$}}}
                                  
\newcommand{\myprob}[1]{\mathop{\mathbb{P}}_{#1}}

\newcommand{\myexp}[1]{\mathop{\mathbb{E}}_{#1}}

\newcommand{\mydelta}[1]{1_{#1}}

\newcommand{\myminimum}[1]{\mathop{\textrm{minimum}}_{#1}}
\newcommand{\mymaximum}[1]{\mathop{\textrm{maximum}}_{#1}}    
\newcommand{\mymin}[1]{\mathop{\textrm{minimize}}_{#1}}
\newcommand{\mymax}[1]{\mathop{\textrm{maximize}}_{#1}}
\newcommand{\mymins}[1]{\mathop{\textrm{min.}}_{#1}}
\newcommand{\mymaxs}[1]{\mathop{\textrm{max.}}_{#1}}  
\newcommand{\myargmin}[1]{\mathop{\textrm{argmin}}_{#1}} 
\newcommand{\myargmax}[1]{\mathop{\textrm{argmax}}_{#1}} 
\newcommand{\myst}{\textrm{s.t. }}

\newcommand{\denselist}{\itemsep -1pt}
\newcommand{\sparselist}{\itemsep 1pt}

\definecolor{pink}{rgb}{0.9,0.5,0.5}
\definecolor{purple}{rgb}{0.5, 0.4, 0.8}   
\definecolor{gray}{rgb}{0.3, 0.3, 0.3}
\definecolor{mygreen}{rgb}{0.2, 0.6, 0.2}

\newcommand{\cyan}[1]{\textcolor{cyan}{#1}}
\newcommand{\red}[1]{\textcolor{red}{#1}}  
\newcommand{\blue}[1]{\textcolor{blue}{#1}}
\newcommand{\magenta}[1]{\textcolor{magenta}{#1}}
\newcommand{\pink}[1]{\textcolor{pink}{#1}}
\newcommand{\green}[1]{\textcolor{green}{#1}} 
\newcommand{\gray}[1]{\textcolor{gray}{#1}}    
\newcommand{\mygreen}[1]{\textcolor{mygreen}{#1}}    
\newcommand{\purple}[1]{\textcolor{purple}{#1}}       

\definecolor{greena}{rgb}{0.4, 0.5, 0.1}
\newcommand{\greena}[1]{\textcolor{greena}{#1}}

\definecolor{bluea}{rgb}{0, 0.4, 0.6}
\newcommand{\bluea}[1]{\textcolor{bluea}{#1}}
\definecolor{reda}{rgb}{0.6, 0.2, 0.1}
\newcommand{\reda}[1]{\textcolor{reda}{#1}}

\def\changemargin#1#2{\list{}{\rightmargin#2\leftmargin#1}\item[]}
\let\endchangemargin=\endlist
                                               
\newcommand{\cm}[1]{}

\newcommand{\mhoai}[1]{{\color{magenta}\textbf{[MH: #1]}}}

\newcommand{\mtodo}[1]{{\color{red}$\blacksquare$\textbf{[TODO: #1]}}}
\newcommand{\myheading}[1]{\vspace{1ex}\noindent \textbf{#1}}
\newcommand{\htimesw}[2]{\mbox{$#1$$\times$$#2$}}

\newcommand{\young}[1]{{\color{blue}$\blacksquare$\textbf{Alternative}: #1}}


\newif\ifshowsolution
\showsolutiontrue

\ifshowsolution  
\newcommand{\Comment}[1]{\paragraph{\bf $\bigstar $ COMMENT:} {\sf #1} \bigskip}
\newcommand{\Solution}[2]{\paragraph{\bf $\bigstar $ SOLUTION:} {\sf #2} }
\newcommand{\Mistake}[2]{\paragraph{\bf $\blacksquare$ COMMON MISTAKE #1:} {\sf #2} \bigskip}
\else
\newcommand{\Solution}[2]{\vspace{#1}}
\fi

\newcommand{\truefalse}{
\begin{enumerate}
	\item True
	\item False
\end{enumerate}
}

\newcommand{\yesno}{
\begin{enumerate}
	\item Yes
	\item No
\end{enumerate}
}

\newcommand{\Sref}[1]{Sec.~\ref{#1}}
\newcommand{\Eref}[1]{Eq.~(\ref{#1})}
\newcommand{\Fref}[1]{Fig.~\ref{#1}}
\newcommand{\Tref}[1]{Table~\ref{#1}}
\title{Gazeformer: Scalable, Effective and Fast Prediction of \\Goal-Directed Human Attention}

\author{Sounak Mondal$^{1}$, Zhibo Yang$^{1,2}$, Seoyoung Ahn$^{1}$, Dimitris Samaras$^{1}$, Gregory Zelinsky$^{1}$, Minh Hoai$^{1,3}$\\
$^{1}$Stony Brook University, \quad $^{2}$Waymo LLC, \quad  $^{3}$VinAI Research
}
\maketitle

\begin{abstract}
   Predicting human gaze is important in Human-Computer Interaction (HCI). 
However, to practically serve HCI applications, gaze prediction models must be scalable, fast, and accurate in their spatial and temporal gaze predictions. Recent scanpath prediction models focus on goal-directed attention (search). Such models are limited in their application due to a common approach relying on trained target detectors for all possible objects, and the availability of human gaze data for their training (both not scalable). In response, we pose a new task called \textit{ZeroGaze}, a new variant of zero-shot learning where gaze is predicted for never-before-searched objects, and we develop a novel model, \textit{Gazeformer}, to solve the ZeroGaze problem. 
 In contrast to existing methods using object detector modules, Gazeformer encodes the target using a natural language model, thus leveraging semantic similarities in scanpath prediction. We use a transformer-based encoder-decoder architecture because transformers are particularly useful for generating contextual representations. Gazeformer surpasses other models by a large margin~(19\%--70\%) on the ZeroGaze setting. It also outperforms existing target-detection models on standard gaze prediction for both target-present and target-absent search tasks. In addition to its improved performance, Gazeformer is more than five times faster than the state-of-the-art target-present visual search model. Code can be found at \href{https://github.com/cvlab-stonybrook/Gazeformer/}{\nolinkurl{https://github.com/cvlab-stonybrook/Gazeformer/}}
\end{abstract}


\begin{figure}[ht!]
\centering
\includegraphics[width=\linewidth]{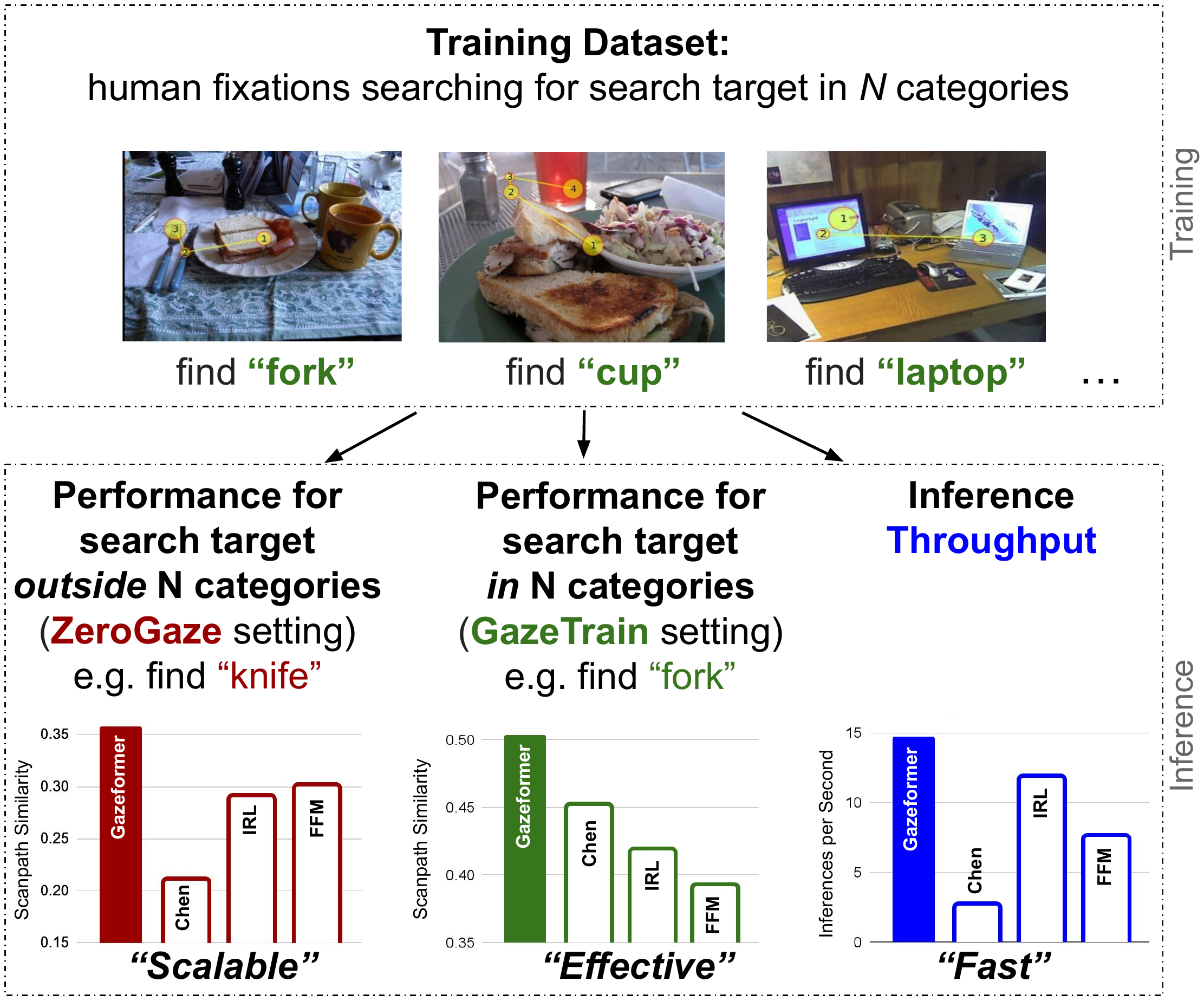}
\caption{An example scenario where human scanpaths for only $N$ categories such as  ``fork'', ``cup'' and ``laptop'' are available during training.  In the traditional \textit{GazeTrain} setting, the model is asked to predict scanpaths for only those $N$ target categories it has seen during training, such as ``fork''. However, in the \textit{ZeroGaze} setting, the model is expected to predict scanpaths for target categories for which training scanpaths are \textit{not available}, such as ``knife''. Gazeformer is overall superior since it is more scalable (better ZeroGaze performance), more effective (better GazeTrain performance) and faster (higher inference speed) than previous methods.}
\label{fig:setting}
\end{figure}
\section{Introduction}
\label{sec:intro}

The prediction of human gaze behavior is a computer vision problem with clear application to the design of more interactive systems that can anticipate a user's attention. In addition to addressing basic 
cognitive science questions, gaze modeling has applications in human-computer interaction (HCI) and Augmented/Virtual Reality (AR/VR) systems~\cite{kapp2021arett, chung2022static, bennett2021assessing}, non-invasive healthcare~\cite{vidal2012wearable, novak2013enhancing}, visual display design~\cite{halverson2007minimal, takahashi2022gaze}, robotics~\cite{kim2021gaze, fujii2018gaze}, education~\cite{chettaoui2023student, beuget2019eye, m_Robello-PERC18},  etc. 
It is likely that gaze modeling will be ubiquitous in all HCI interfaces in the near future. Hence, there is a need for gaze prediction models that are reliable and effective yet efficient and capable of fast inference for use with edge devices (e.g., smartglasses,  AR/VR headsets).

Our focus is on goal-directed behavior (not free viewing), and specifically the prediction of gaze fixations as a person searches for a given target-object category (e.g., a clock or fork). Visual search is engaged in innumerable task-oriented human activities (e.g., driving) in both real and AR/VR environments. The gaze prediction problem for visual search takes an image input and outputs a sequence of eye movements conditioned on the target. 
Existing architectures for search fixation prediction use either panoptic segmentation maps \cite{yang2020predicting} or object detection modules \cite{chen2021predicting, yang2022target} to encode specific search targets. However, the applicability of these methods is essentially limited to the relatively small number of target objects for which panoptic segmentation or object detector models can be realistically trained. For example, to predict the fixations in search for ``pizza cutter'', which is an object category not included in the dataset used to train the backbone detector in \cite{chen2021predicting}, a new ``pizza cutter'' detector would need to be trained, meaning this and related methods do not scale beyond their training data. Relatedly, existing approaches~\cite{yang2020predicting, chen2021predicting, yang2022target} have required the laborious collection of large-scale datasets of search behavior for each target category ($\sim3000$ scanpaths per category) included in the dataset, an approach that is unscalable to the innumerable potential targets that humans search for in the wild.
 
We therefore introduce a new problem that we refer to as \textit{ZeroGaze}, which is an extension of zero-shot learning to the gaze prediction problem. Under ZeroGaze, a model must predict the fixations that a person makes while searching for a target (e.g., a fork) despite the unavailability of search fixations (e.g., a person looking for a fork) for model training. A more challenging version of the problem is when there are no trained detectors for the target (e.g., a fork detector) either.  
This is in contrast to the traditional ``GazeTrain'' setting (where the model has been trained on gaze behavior for all the categories that it encounters during inference). 
To address the ZeroGaze problem, we propose \textit{Gazeformer}, a novel multimodal model for scanpath prediction. Gazeformer is scalable because it does not require training a backbone detector on the fixation behavior of people searching for specific object categories. 
We use a linguistic embedding of the target object from a language model (RoBERTa~\cite{liu2019roberta}) and a transformer encoder-decoder architecture \cite{NIPS2017_3f5ee243} to model the joint image-target representations and the spatio-temporal context embedded within it. Given that recent language models can encode any object using any text string, they provide a powerful and scalable solution to the representation of target categories for which no gaze data is available for training. Additionally, Gazeformer decodes the fixation time-steps in \textit{parallel} using a transformer decoder, providing significantly faster inference than any other method, which is necessary for gaze tracking applications in HCI products (especially in wearable edge devices). We show the advantages of our model in Fig. \ref{fig:setting}. The specific contributions of this study are:

\begin{enumerate}\denselist
    \item We introduce the \textit{ZeroGaze} task, where a model must predict the search fixation scanpath for a new target category without training on prior knowledge of the gaze behavior to that target. 
    
    \item We devise a novel multimodal transformer-based architecture called \textit{Gazeformer}, which learns the interaction between the image input and the language-based semantic features of the target. 
      \item We propose a novel and effective scheme of fixation modeling that uses a set of Gaussian distributions in continuous image space rather than learning multinomial probability distributions over patches, resulting in an intuitive distance-based objective function. 

      \item We achieve a new state-of-the-art in search scanpath prediction. It outperforms baselines, not only in our new ZeroGaze setting, but also in the traditional setting where models are trained on the gaze behavior (GazeTrain) and in target-absent search. \textit{Gazeformer} also generalizes to unknown categories while being up to more than 5 times faster than previous methods.

    \end{enumerate}



\section{Related Work}

\myheading{Gaze Prediction for Search Task}.
Beginning with~\cite{IttiPAMI98}, most efforts for gaze prediction have been focused on using saliency maps \cite{borji2013state, kummerer2017understanding, kruthiventi2017deepfix, huang2015salicon, kummerer2014deep} to predict free-viewing behavior.
In free-viewing, attention is driven entirely by features extracted from the visual input, whereas in a search task, a top-down goal is used as well to guide attention. 
An early study predicted search fixations for two target categories, microwave ovens and clocks, in the identical scene contexts\cite{zelinsky2019benchmarking,zelinsky2021predicting}. Inverse Reinforcement Learning was used to predict search fixations in \cite{yang2020predicting}, exploiting a recent large-scale dataset of search fixations encompassing 18 targets categories (COCO-Search18 \cite{chen2021coco}). \cite{chen2021predicting} proposed a model that predicts scanpaths using task-specific attention maps and performs well in several visual tasks including visual search. \cite{chen2022characterizing, yang2022target} addressed the target-absent problem. 

\myheading{Zero Shot Learning}.
In machine learning, Zero Shot Learning refers to a problem whereby samples encountered during inference time are from classes that have not been observed during training. Zero Shot Learning has been explored in the context of image classification~\cite{elhoseiny2013write}, object detection~\cite{rahman2018zero} and many more computer vision tasks, and Zhang \etal~\cite{zhang2018finding} extended it to zero-shot visual search by proposing a network that predicts scanpaths for unseen novel objects. However, their model requires seeing a \textit{visual image of an exemplar of the target class} as input. 
In contrast, we focus on the problem of ``ZeroGaze'' prediction, where the model must infer scanpaths for targets without relying on template matching or fixation data specific to the target category. 

 \begin{figure*}[ht!]
\centering
\includegraphics[width=\linewidth]{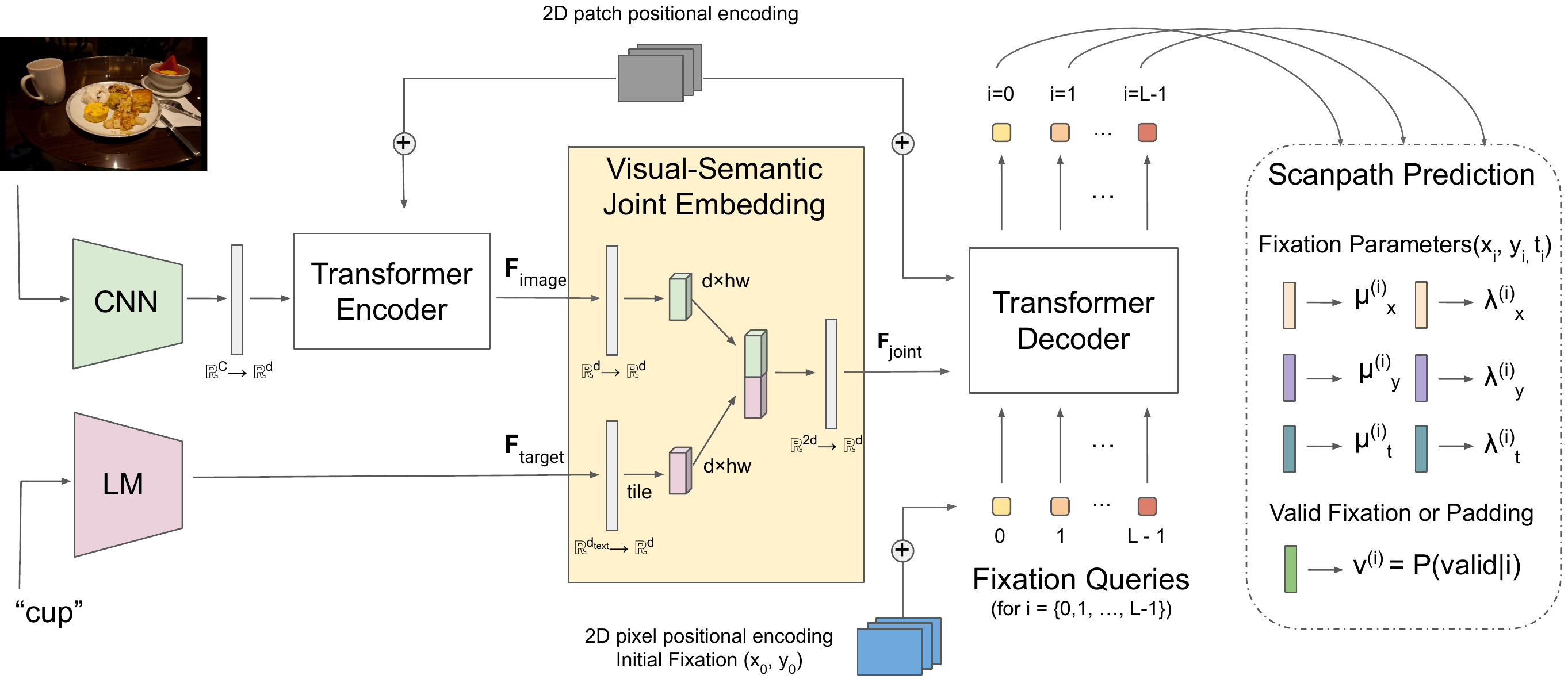}
\caption{{\bf Overall architecture of the Gazeformer model}. A transformer encoder block is used to contextualize CNN-extracted visual features. After being jointly embedded with target semantic features from a language model (LM), the resultant features interact with $L$ fixation queries in a  transformer decoder. The $L$ output encodings are passed to 7 MLP layers to obtain coordinates, duration and validity for all possible fixations in parallel. We use ResNet-50~\cite{he2016deep} as the CNN and RoBERTa~\cite{liu2019roberta} as the LM for our experiments. }
\label{fig:arch}
\end{figure*}
\myheading{Transformers}.
Transformers~\cite{NIPS2017_3f5ee243} use an attention mechanism to provide context for any position in the input sequence. 
Recently, transformers have made considerable impact in Computer Vision\cite{dosovitskiy2021an, touvron2021training, liu2021swin}. 
A particular line of transformer-based models (DETR~\cite{carion2020end}, MDETR~\cite{kamath2021mdetr}) use a convolutional back-end to extract local visual features, which are then used in a transformer encoder-decoder architecture to perform contextual reasoning. 
A recent work~\cite{chao2021transformer} predicts free-viewing viewport scanpath for 360$^{\circ}$ videos with a transformer encoder based architecture.
To our best knowledge, ours is the first study that applies a \textit{transformer encoder-decoder} architecture with \textit{multimodal understanding} to predict human scanpaths for \textit{visual search}.

\section{Gazeformer}
Two factors enable Gazeformer to generalize to the ZeroGaze scanpath prediction problem. 
First, Gazeformer jointly embeds visual and semantic features of a search target, with the latter extracted from a language model (RoBERTa\cite{liu2019roberta}). This means that the target is encoded as an interaction between image features and real-world semantic relationships.  
Second, the self-attention mechanism of the encoder allows it to learn the global and local context of the scene that previous convolution-based methods missed. Gazeformer learns where targets are usually located, meaning that it learns object co-occurrence and other object-scene relationships that convey information about target location (e.g., first finding a kitchen counter top when searching for a microwave), and this enables it to generalize well to target-absent search, where scene context is crucial. 
Two additional factors improve Gazeformer's scanpath prediction capabilities. First, our model uses a transformer-based decoder to predict an entire sequence of search fixations in parallel (including when to terminate). This parallel decoding not only offers a considerable speed up in inference time for scanpath prediction, but also allows the model to learn long-range dependencies between fixations in a bidirectional manner, differing from the standard unidirectional sequential scanpath prediction approach\cite{yang2020predicting, chen2021predicting, yang2022target}. 
Second, instead of performing a patch-wise prediction for generating a sequence of fixations, as in prior approaches~\cite{yang2020predicting, chen2021predicting, yang2022target}, we directly regress the fixation locations parameterized by Gaussian distributions. This enables Gazeformer to learn an effective distance-based objective function for predicting the entire sequence of scanpath fixations at once. 

\subsection{Architecture}
The overview of Gazeformer architecture is provided in \Fref{fig:arch}. First, image features are extracted from a ResNet-50 backbone and further processed through a transformer encoder to obtain contextual image features $\F_{image}$. The semantic embedding $\F_{target}$ of a search target is extracted from RoBERTa. Consequently, we compute $\F_{joint}$, a joint visual-semantic embedding of image and target features. Given the features $\F_{joint}$ and learnable fixation queries (which represent the time step information of each fixation), a transformer decoder produces a sequence of fixation embeddings \textit{in parallel} for each time-step. The fixation embeddings are then processed through seven independent Multi-Layer Perceptron (MLP) layers, which yield (1) fixation coordinates, (2) fixation duration, and (3) scanpath termination information as final model output. Below are the details of each model component.


\myheading{Image Feature Encoding.} We extract features for an image by resizing it to $1024{\times}640$ resolution and passing it through a ResNet-50~\cite{he2016deep} network to extract a feature map of dimensions ${C{\times} h {\times} w}$, where $C=2048, h=20, w=32$.  We intentionally choose the resize resolution to be  $1024{\times}640$ so that $h=20$ and $w=32$, in order to have the same input granularity as baseline representations \cite{yang2020predicting, yang2022target}. We flatten  the spatial dimensions of this feature map to obtain a 2D feature tensor of size $C{\times} hw$. 
Since $C=2048$ is prohibitively large for further computation, we use a linear layer to reduce the number of feature dimensions from $2048$ to a smaller value $d$. This forms the input to a transformer encoder block that consists of a cascade of $N_{enc}$ standard transformer encoder~\cite{NIPS2017_3f5ee243} layers. Each layer uses multi-head self-attention, feed-forward networks, and layer normalization to find contextual embeddings of the input. To indicate the location of each patch, we use a fixed sinusoidal $2D$ positional encoding as in \cite{carion2020end}. This encoder block creates a task-agnostic contextualized representation of the image in the form of feature tensor $\F_{image}\in \mathbb{R}^{d{\times}hw}$. 


\myheading{Target Feature Semantic Encoding.} To extract target features, we use the language model RoBERTa~\cite{liu2019roberta} (we use the RoBERTa-base variant) to encode the text string signifying the target category (such as ``potted plant'') as a tensor $\mathbf{F}_{target}\in\mathbb{R}^{d_{text}}$, where $d_{text}=768$. Unlike word embedding frameworks like Word2vec~\cite{Mikolov2013EfficientEO} and GloVe~\cite{pennington2014glove}, RoBERTa can encode both single-word category names such as ``car'' and multi-word target category names such as ``stop sign'' to a fixed length  vector of dimension ${d_{text}}$.

\myheading{Joint Embedding of Image and Target Feature.}
To create a joint visual-semantic embedding space of image and target features, we first independently map $\mathbf{F}_{image}$ and $\mathbf{F}_{target}$ to a shared multimodal $d$-dimensional latent space using modality-specific linear transformations. Then we tile the modality-specific transformation of $\mathbf{F}_{target}$ spatially $hw$ times. We concatenate the transformed image features and the transformed and tiled target features (both now sized ${d{\times}hw}$) along the channel dimension and obtain a 2D tensor of size  ${2d{\times}hw}$, which is again linearly projected (with ReLU activation) to the dimension size of~$d$. The final jointly embedded visual-semantic features are $\mathbf{F}_{joint}\in \mathbb{R}^{d\times h w}$. This joint embedding step differs from previous multimodal methods \cite{kamath2021mdetr, yang2022tubedetr} in that we append semantic features of a search target only after a target-agnostic contextual image representation is obtained from the encoder.

\myheading{Fixation Decoding.} We decode an entire sequence of fixations in \textit{one go} using a transformer decoder. Like most transformer architectures, we have a maximum output scanpath sequence length $L$ which requires fixation sequences to be \textit{padded} if their lengths are smaller than $L$. The decoder contains $N_{dec}$ standard transformer decoder layers~\cite{NIPS2017_3f5ee243} (stacked self-attention and cross-attention layers with minor modifications) and processes $\mathbf{F}_{joint}$ along with a set of~$L$ \textit{fixation queries} $Q_i\in \mathbb{R}^d, i\in\{0,1,...,L-1\}$. 
The fixation queries are randomly initialized learnable embeddings that provide fixation time step information. At each decoder layer, the latent fixation embeddings interact with each other through self-attention, and also interact with $\mathbf{F}_{joint}$ through encoder-decoder cross-attention. Note that before each cross-attention step, a fixed 2D positional encoding is added to $\mathbf{F}_{joint}$ in order to provide positional information about the patches. We encode initial fixation location $(x_0, y_0)$ (which affects the generated scanpath) using another fixed 2D positional encoding and add it to $Q_{0}$. 
The output of the transformer decoder block is $\mathbf{F}_{dec}\in \mathbb{R}^{d{\times} L}$.

\myheading{Scanpath Prediction.} Previous scanpath prediction models  \cite{yang2020predicting, chen2021predicting, yang2022target} predict fixations on a patch-level granularity by reducing the image space into a set of discrete locations and generating a multinomial probability distribution over them. However, one limitation of this approach is that all patches are at the same distance from each other---patches closer to each other are treated the same way as patches that are further apart. To remedy this, we propose to directly \textit{regress} the raw fixation co-ordinates for each $L$ possible time steps from $\mathbf{F}_{dec}$. To incorporate the inter-subject variability 
in human fixations, we model fixation locations and durations using Gaussian distributions, and consequently regress the mean and log-variance of 2D co-ordinates and duration of a fixation  using six separate MLP layers. For the $i^{th}$ fixation, let $\mu_{x_i}, \mu_{y_i}, \mu_{t_i}, \lambda_{x_i}, \lambda_{y_i},  \lambda_{t_i}$ denote the mean and log-variance for the $x$ and $y$ positions and duration $t$. Using the reparameterization trick~\cite{kingma2013auto}, the fixation co-ordinates $x_i$ and $y_i$ and duration $t_i$ are estimated as follows:
\begin{align}
 &   x_i = \mu_{x_{i}} + \epsilon_{x_{i}} {\cdot} \exp({0.5 \lambda_{x_{i}}}), \quad 
    y_i = \mu_{y_{i}} + \epsilon_{y_{i}} {\cdot} \exp({0.5 \lambda_{y_{i}}}), \nonumber  \\
&    t_i = \mu_{t_{i}} + \epsilon_{t_{i}} {\cdot} \exp({0.5 \lambda_{t_{i}}}), \quad \epsilon_{x_{i}}, \epsilon_{y_{i}}, \epsilon_{t_{i}} \in \mathcal{N}(0,1).
\end{align}
This allows our network to be fully differentiable despite having a probabilistic component. Finally, since we use padding, we make a prediction for each output if the current step belongs to a valid fixation or a padding token. This is done by an MLP classifier with softmax activation which classifies each of the $L$ slices of $\mathbf{F}_{dec}$ to be a valid fixation or padding token. During inference, we collect the sequence of $(x_i, y_i, t_i, v_i)$ ordered quads for $i \in \{0,1,...,L-1\}$, where $x_i$, $y_i$ are the coordinates of fixation $i$, $t_i$ is the duration of fixation $i$, and $v_i$ the probability of being a valid fixation. We traverse the sequence from $i=0$ to $i=L-1$ and terminate the sequence at the first padding token ($v_i< 0.5$). We experimentally confirmed that our model performs better with regression than with classification. Architecture details are in the supplemental.

\subsection{Training}

We view scanpath prediction as a sequence modeling task. The total multitask loss $\mathcal{L}$ to train our network with backpropagation for a minibatch of $M$ samples is:

\begin{align}
    &\mathcal{L} = \frac{1}{M} \sum_{j=1}^{M} \left( \mathcal{L}_{xyt}^{j} + \mathcal{L}_{val}^{j} \right), \\
    \textrm{where } &\mathcal{L}_{xyt}^{j} = \frac{1}{l^{j}}  \sum_{i=0}^{l^{j}-1} \left( |x_i^{j}  - \hat{x}_i^{j}| + |y_i^{j}  - \hat{y}_i^{j}| + |t_i^{j}  - \hat{t}_i^{j}| \right), \nonumber \\ 
    &\mathcal{L}_{val}^{j} = - \frac{1}{L}  \sum_{i=0}^{L-1} \left(v_i^{j} \log \hat{v}_i^{j} + (1-v_i^{j})\log (1 - \hat{v}_i^{j})\right). \nonumber 
\end{align}
Here $s^j = \{(x_{i}^{j}, y_{i}^{j}, t_{i}^{j})\}_{i=0}^{L-1}$ is the  predicted scanpath and $L$ is the maximum scanpath length. $l^{j}$ is the length of the ground truth scanpath $\hat{s}^j = \{(\hat{x}_{i}^{j}, \hat{y}_{i}^{j}, \hat{t}_{i}^{j})\}_{i=0}^{l^{j}-1}$. $\hat{v}_i^{j}$ is a binary scalar  representing ground truth of $i^{th}$ token in $s^j$ being a valid fixation or padding and $v_i^{j}$ is the probability of that token being a valid fixation as estimated by our model. 
The losses included in $\mathcal{L}_{xyt}$ are the $L_1$ regression losses for $x$ and $y$ co-ordinates and duration $t$ of the fixations. $\mathcal{L}_{val}$ is the negative log likelihood loss for validity prediction for each token. We find the $L_1$ loss to be more appropriate for scanpath prediction because of the intrinsic variability in human fixation locations in a multi-subject setting.  Note that we mask out ground truth padded tokens while calculating $\mathcal{L}_{xyt}$ to only account for valid fixations in ground truth. Optimization and training details are in the supplemental.

%

\section{Experiments}
We evaluate and compare the model performance under two prediction scenarios: ZeroGaze and GazeTrain. In the ZeroGaze setting, a model must predict search fixations for a category not encountered during training. GazeTrain is the traditional scanpath prediction setting where a model is trained and tested to predict search fixations using the same set of target categories. All experiments pertaining to ZeroGaze and GazeTrain settings were conducted using the target-present data of the COCO-Search18~\cite{chen2021coco} dataset following the evaluation scheme in~\cite{yang2020predicting}. COCO-Search18 consists of 3101 target-present (TP) images, and 100,232 scanpaths collected for 18 categories from 10 subjects. We use the original train-validation-test split provided by \cite{yang2020predicting} and report all results on the test set. We also demonstrate the superior generalizability of our method by comparing it with other baselines on the target-absent data of COCO-Search18 after training the model on either (1) the target-present data or (2) the target-absent data. The IRL model from Yang \etal~\cite{yang2020predicting}, the model from Chen \etal~\cite{chen2021predicting} and a recent FFM model from Yang \etal~\cite{yang2022target} are used as baseline methods. Since only Chen \etal~\cite{chen2021predicting}'s model predict duration, we also provide a variation of Gazeformer called \textit{Gazeformer-noDur} which does not train on or predict fixation duration for fair comparison with the IRL and FFM baselines. We remap the predicted fixations from Chen \etal~\cite{chen2021predicting}'s model to a $20{\times}32$ grid following \cite{yang2022target}. All of the models except IRL predict search stopping behavior, and we use the baselines' stopping prediction modules whenever available to obtain a fairer comparison. We try to maintain the same hyperparameters across settings for all models to evaluate generalizability and scalability.

\subsection{Implementation Details}

Unless explicitly specified otherwise, the results reported in this paper are for the Gazeformer model with 6 encoder layers, 6 decoder layers, 8 attention heads per multihead attention block, and hidden size ($d$) of 512.  Following \cite{yang2020predicting}, we set maximum scanpath length to 7 (including the initial fixation) for our experiments. 
Due to the limited availability of training data and computational resources for additionally finetuning the backbone networks, we 
use frozen ResNet-50 and RoBERTa encodings. Models generate 10 scanpaths per test image by sampling at each fixation, and the reported metrics are averages. 

\subsection{Metrics}
We report model performance in search scanpath prediction using a diverse set of metrics. \textit{Sequence Score (SS)} \cite{yang2020predicting} converts scanpaths into strings of fixation cluster IDs and a string matching algorithm \cite{needleman1970general} measures similarity between two strings. 
\textit{Semantic Sequence Score (SemSS)}~\cite{yang2022target} modifies SS by converting scanpaths to strings of fixated objects in the scene instead of cluster IDs, increasing interpretability. Unlike the implementation of SemSS by \cite{yang2022target} that disregarded the COCO stuff classes, we include both thing and stuff classes for object assignment. We also include \textit{Fixation Edit Distance (FED)} and \textit{Semantic Fixation Edit Distance (SemFED)}, which convert scanpaths to strings like the SS and SemSS metrics, respectively, but use the Levenshtein algorithm\cite{Levenshtein1965BinaryCC} to measure scanpath dissimilarity. \textit{Multimatch (MM)} \cite{anderson2015comparison, dewhurst2012depends} is a popular metric that measures the scanpath similarity at the pixel level. 
Here, MM indicates the average of the shape, direction, length, and position scores (individual metrics in supplemental). 
\textit{Correlation Coefficient (CC)~\cite{jost2005assessing}} measures the correlation between the normalized model and the human fixation map (i.e., the 2D distribution map of all fixations convolved with a Gaussian). \textit{Normalized Scanpath Saliency (NSS)~\cite{peters2005components}} is a discrete approximation of CC, which averages the values of a model's fixation map from the human fixated locations~\cite{bylinskii2018different}. Higher values for SS, SemSS, MM, NSS and CC indicate greater similarity between model-generated and human search scanpaths. The direction is opposite for FED and SemFED. We include SS, FED, SemSS and SemFED both with duration (as in~\cite{cristino2010scanmatch}) and without duration. \footnote{We have updated the implementation of SemSS and SemFED metrics. Please find the updated SemSS and SemFED scores corresponding to the Tables~\ref{table:all_results} and~\ref{table:ta_results} in the appendix (Sec.~\ref{sec:appendix}).}

\setlength{\tabcolsep}{4pt}
\begin{table*}[ht!]
\centering
\begin{tabular}{l|cc|cc|cc|cc|c|c|c}
\toprule 
 & \multicolumn{2}{c|}{SS$\bm{\uparrow}$} & \multicolumn{2}{c|}{SemSS$\bm{\uparrow}$} & \multicolumn{2}{c|}{FED$\bm{\downarrow}$} & \multicolumn{2}{c|}{SemFED $\bm{\downarrow}$} & MM  & CC & NSS\\  
 & w/o Dur & w/ Dur & w/o Dur & w/ Dur & w/o Dur & w/ Dur & w/o Dur & w/ Dur & $\bm{\uparrow}$ & $\bm{\uparrow}$ & $\bm{\uparrow}$
 \\
 \Xhline{0.75pt}
IRL~\cite{yang2020predicting} & 0.290 & - & 0.314 & - & 4.606 & - & 4.377 & - & 0.774 & 0.241 & 4.018 \\
Chen \etal~\cite{chen2021predicting} & 0.210 & 0.041 & 0.211 & 0.034 & 5.720 & 210.498 & 5.608 & 211.636 &0.717 & 0.002 & 0.001 \\
FFM~\cite{yang2022target} & 0.300 & - & 0.334 & - & 3.271 & - & 2.918 & - & 0.731 & 0.271 & \textbf{5.247}\\ \hline
Gazeformer-noDur & \textbf{0.359} & - & \textbf{0.391} & - & 2.788 & - & 2.474 & - &\textbf{0.822} & 0.316 & 4.671 \\
Gazeformer & 0.358 & \textbf{0.312} & \textbf{0.391} & \textbf{0.348} & \textbf{2.766} & \textbf{12.505} & \textbf{2.438} & \textbf{10.391} & 0.812 & \textbf{0.324} & 4.929\\
\bottomrule 
\end{tabular}\\
(a)\\

\begin{tabular}{l|cc|cc|cc|cc|c|c|c}
\toprule 
 & \multicolumn{2}{c|}{SS$\bm{\uparrow}$} & \multicolumn{2}{c|}{SemSS$\bm{\uparrow}$} & \multicolumn{2}{c|}{FED$\bm{\downarrow}$} & \multicolumn{2}{c|}{SemFED $\bm{\downarrow}$} & MM & CC & NSS\\  
 & w/o Dur & w/ Dur & w/o Dur & w/ Dur & w/o Dur & w/ Dur & w/o Dur & w/ Dur & $\bm{\uparrow}$ & $\bm{\uparrow}$ & $\bm{\uparrow}$
 \\
 \Xhline{0.75pt}
Human & 0.490 & 0.409 & 0.548&0.456 & 2.531&11.526 & 1.637&8.086 & 0.857 & 0.472 & 8.129\\ \hline
IRL~\cite{yang2020predicting} & 0.418&- & 0.499&- & 2.722&- & 2.182&- & 0.833 & 0.434 & 6.895\\
Chen \etal~\cite{chen2021predicting} & 0.451&0.403 & 0.504&0.446 & \underline{2.187} &\underline{10.795} & 1.788&8.782 & 0.820 & \underline{0.547} & 6.901 \\
FFM~\cite{yang2022target} & 0.392&- & 0.443&- & 2.693&- & 2.284&- & 0.808 & 0.370 & 5.576 \\ \hline
Gazeformer-noDur & \underline{\textbf{0.504}}&- & \textbf{0.534}&- & \underline{\textbf{2.061}}&- & \textbf{1.742}&- & 0.849  & \underline{0.559} & \underline{8.356}\\
Gazeformer & \underline{\textbf{0.504}}&\underline{\textbf{0.451}} & 0.525&\underline{\textbf{0.485}} & \underline{2.072}& \underline{\textbf{9.708}} & 1.810 &\underline{\textbf{7.688}} & \textbf{0.852} & \underline{\textbf{0.561}}& \underline{\textbf{8.375}} \\
\bottomrule 
\end{tabular}\\
(b)
\caption{Model performance comparison under (a) ZeroGaze setting, (b) traditional GazeTrain setting.  Best performance is highlighted in bold. Performances that exceed human consistency are underlined.}
\label{table:all_results}
\end{table*}

\begin{figure*}[ht!]
\centering
\includegraphics[width=\linewidth]{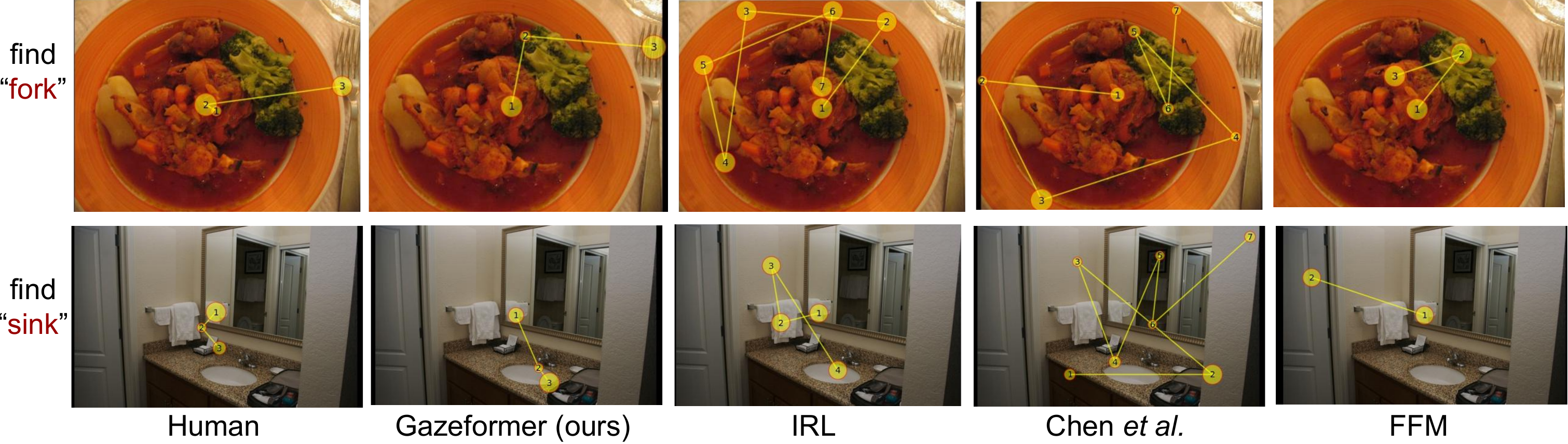}
\caption{Visualization of scanpath prediction under the ZeroGaze setting. 
The number and radius indicate the fixation
order and duration (if predicted), respectively. Our Gazeformer model predicts efficient, human-like scanpaths.}
\label{fig:zerogaze}
\end{figure*}

\subsection{ZeroGaze Setting}

Models were trained on 17 of the COCO-Search18 categories and tested on the one left-out category in a cross-validation manner. For example, to evaluate ZeroGaze performance on category $\mC$, we remove the category $\mC$ scanpaths from the training data and predict gaze behavior for this category during testing. Thus, we ensure that a model does not see any behavioral data related to category $\mC$ for ZeroGaze prediction. Performance was averaged over all categories and weighted by the number of test cases for each category. As shown in \Tref{table:all_results}(a), Gazeformer achieves state-of-the-art ZeroGaze performance, outperforming the baselines by considerable margins across almost all metrics. See Fig.~\ref{fig:zerogaze} for some qualitative results. More qualitative results can be found in the supplemental.


\subsection{Traditional GazeTrain Setting}

In the traditional GazeTrain setting, models were trained with target-present search fixations on 18 different target categories (from the COCO-Search18 dataset) and tested on the same set of categories but with unseen test cases. We compared the predictive performance of Gazeformer and previous models of search scanpath prediction. Table \ref{table:all_results}(b) shows that Gazeformer generally outperforms the baselines (by significant margins for some metrics) and is at (or slightly exceeds) a noise ceiling imposed by human consistency rates (i.e., predictions cannot be much better). Hence, Gazeformer is the new state-of-the-art in target-present visual search scanpath prediction. We include qualitative results of Gazeformer and other baseline models in the supplemental.



\setlength{\tabcolsep}{5pt}
\begin{table*}[ht!]
\centering

\begin{tabular}{l|cc|cc|c|cc|cc|c}
\toprule 
& \multicolumn{5}{|c|}{Trained on Target-Present} & \multicolumn{5}{c}{Trained on Target-Absent}
\\
 \Xhline{0.75pt}
 & \multicolumn{2}{c|}{SS$\bm{\uparrow}$} & \multicolumn{2}{c|}{SemSS$\bm{\uparrow}$} & MM &\multicolumn{2}{c|}{SS$\bm{\uparrow}$} & \multicolumn{2}{c|}{SemSS$\bm{\uparrow}$} & {MM}\\
 & w/o Dur & w/ Dur & w/o Dur & w/ Dur & $\bm{\uparrow}$ & w/o Dur & w/ Dur & w/o Dur & w/ Dur & $\bm{\uparrow}$
 \\
 \Xhline{0.75pt}
Human & 0.398 & 0.369 & 0.436 & 0.404 & 0.838 & 0.398 & 0.369 & 0.436 & 0.404 & 0.838\\ \hline
IRL~\cite{yang2020predicting} & 0.304 & - & 0.349 & - & 0.808 & 0.323 & - & 0.378 & - & 0.805 \\
Chen \etal~\cite{chen2021predicting} & 0.350 & 0.330 & 0.395 & 0.380  & 0.813 & 0.345 & 0.323 & 0.347 & 0.335 & 0.799\\
FFM~\cite{yang2022target} & 0.360 & - & 0.413 & -  & 0.814 & 0.362 & - & 0.413 & - & 0.814\\ \hline
Gazeformer-noDur & 0.366 & - & \textbf{0.419} & -  & \textbf{0.833} & 0.369 & - & 0.422 & - & 0.830 \\
Gazeformer & \textbf{0.368} & \textbf{0.356} & \textbf{0.419} & \textbf{0.399}
 & 0.825 & \textbf{0.375}&\textbf{0.361} & \underline{\textbf{0.438}} & \underline{\textbf{0.417}}& \underline{\textbf{0.844}}\\
\bottomrule 
\end{tabular}
\caption{Performance comparison for models tested on target-absent data after training on target-present data and target-absent data. Best performance is highlighted in bold. Performances that exceed human consistency are underlined.}
\label{table:ta_results}
\end{table*}

\subsection{Generalizability to Target-Absent Search}

To demonstrate generalizability, we evaluated model performance on COCO-Search18 target-absent trials under two settings - (1) where the model is trained on only target-present data, and (2) where the model is trained on only target-absent data. 
The first setting should be direct evidence for the capability of the model to learn the important contextual relationships between scene objects that humans use to guide their search for a target when it is absent from the scene. 
The second setting tests the model's capability to learn search behavior for the target-absent task by training on target-absent human trials. 
We report the results for both settings in Table~\ref{table:ta_results}.  Similar to the target-present results, Gazeformer again outperforms all baselines in multiple metrics, establishing new SOTA in target-absent scanpath prediction as well (although the human noise ceiling was only achieved with training on the target-absent data). Gazeformer appears to be learning scene context and object relationships well and generalizes to target-absent search without any change in architecture or hyperparameters. We include a comprehensive version of Table \ref{table:ta_results} and qualitative evidence for Gazeformer's use of context through attention maps in the supplemental.

\subsection{Inference Speed}

\begin{table}[ht!]
\setlength{\tabcolsep}{3pt}
\centering
\begin{tabular}{lccc}
\toprule 
 & Time (in ms)$\bm{\downarrow}$  & Inferences/s $\bm{\uparrow}$ & Speedup $\bm{\uparrow}$\\  
 \Xhline{0.75pt}
 Chen \etal~\cite{chen2021predicting} & 386 & 2.59 & 1X  \\

 FFM~\cite{yang2022target} & 133 & 7.52 & 2.9X\\
 IRL~\cite{yang2020predicting} & 85 & 11.77 & 4.5X\\
 Gazeformer & \textbf{68} &\textbf{14.71} & \textbf{5.7X}\\
 \bottomrule 
 \end{tabular}
 \caption{Comparison of inference speeds. Gazeformer achieves 5.7X speedup over highly competent Chen \etal's~\cite{chen2021predicting} model.}
\label{table:speed_results}
\end{table}

 Despite clearly outperforming baselines in search scanpath prediction, Gazeformer is many times faster. We attribute this to the parallel decoding mechanism in the transformer decoder that predicts the entire scanpath in one go while other baselines are sequential in nature. We report the average time measured on COCO-Search18's test split of target-present trials in Table 
\ref{table:speed_results}. 
We measure time in the real-world setting where the model receives and operates on one search task case (image-target pair) at a time. 

\subsection{Extension to Unknown Categories}

\def\subFigSz{0.32\linewidth}
\begin{figure}[ht!]
\centering
  \includegraphics[width=\subFigSz]{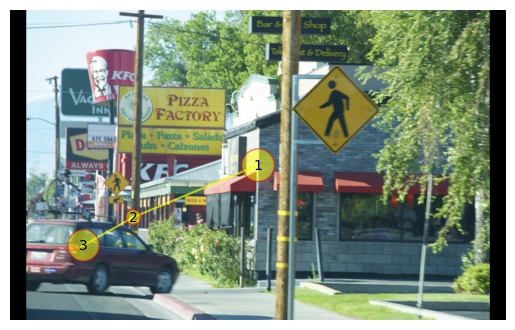} 
  \includegraphics[width=\subFigSz]{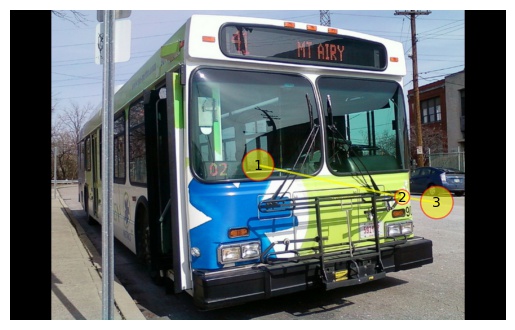}  
  \includegraphics[width=\subFigSz]{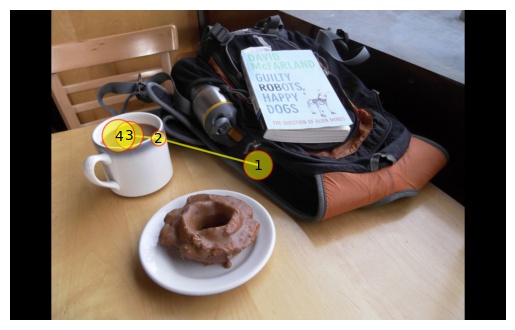}\\
  \makebox[\subFigSz]{\small{find ``hatchback''}}
\makebox[\subFigSz]{\small{find ``sedan''}}
\makebox[\subFigSz]{\small{find ``mug''}}\\

\includegraphics[width=\subFigSz]{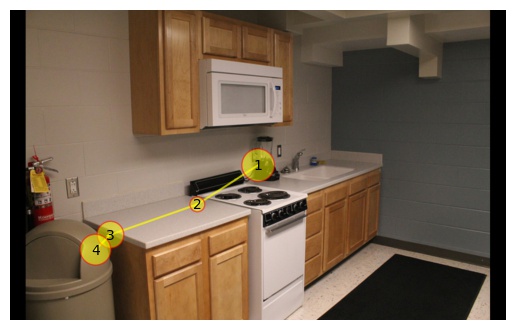} 
  \includegraphics[width=\subFigSz]{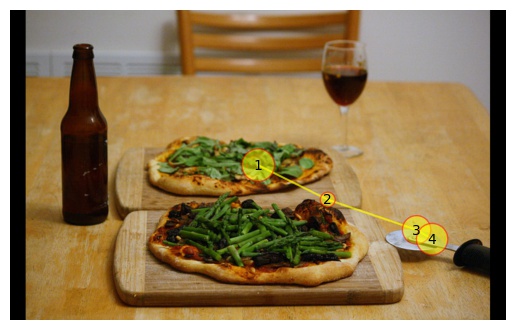}  
  \includegraphics[width=\subFigSz]{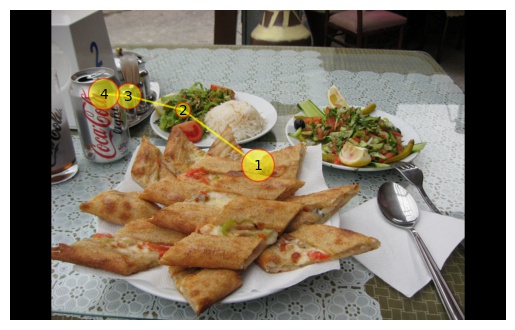}\\
  \makebox[\subFigSz]{\small{find ``trash can''}}
\makebox[\subFigSz]{\small{find ``pizza cutter''}}
\makebox[\subFigSz]{\small{find ``soda can''}}
\caption{Generalization of Gazeformer to unknown categories. Top row shows extensions to non-canonical names of COCO-Search18's categories; bottom row shows extensions beyond COCO-annotated categories.} 
\label{fig:lm_extends}
\end{figure}

Since the proposed model uses the RoBERTa language model to encode target objects, hypothetically it can generalize to searching for any object that can be described in words or text.  First, we generate and visualize new scanpaths using the synonyms or hyponyms of the COCO-Search 18 objects to define targets (e.g., replacing ``cup'' with ``mug'' and ``car'' with ``hatchback'' or ``sedan''). We also investigate if our model can extend beyond the MS-COCO dataset used to train backbone models, to completely unseen targets like ``trash can'', ``pizza cutter'' or ``soda can''
. Quantitative comparison is not possible because there are no human search fixation data, so we only generate and visualize scanpaths for these cases. As shown in Fig.~\ref{fig:lm_extends}, Gazeformer generates plausible natural-looking scanpaths that successfully find the unknown target. 
Previous methods are confined to a predefined set of categories due to their reliance on detector/panoptic maps and fail to handle these scenarios.


\subsection{Impact of Language Embeddings}

Gazeformer uses RoBERTa\cite{liu2019roberta} linguistic embeddings as target representations, and to explore their role in model performance, we replaced the 18 target category embeddings from RoBERTa with fixed random embeddings. 

\setlength{\tabcolsep}{8pt}
\begin{table}[ht!]
\centering

\begin{tabular}{lccc}
\toprule 
Target Embedding & SS$\bm{\uparrow}$ & FED$\bm{\downarrow}$& NSS$\bm{\uparrow}$\\  
\midrule 
RoBERTa & \textbf{0.359} & \textbf{2.788} & \textbf{4.671}\\
Fixed Random & 0.336 & 2.873 & 4.524\\
\bottomrule 
\end{tabular}\\
(a)\\[0.2cm]

\setlength{\tabcolsep}{5pt}
\begin{tabular}{llccc}
\toprule 
Category & Embedding & SS$\bm{\uparrow}$ & FED$\bm{\downarrow}$& NSS$\bm{\uparrow}$\\  
\midrule 
\multirow{2}{1.4cm}{stop sign} & RoBERTa & 0.430 & 2.256 & 4.559\\
& Random &  0.358 & 2.176 & 4.258\\\hline
\multirow{2}{1.4cm}{clock} & RoBERTa & 0.399 & 2.600 & 2.110\\
& Random &  0.310 & 3.274 & 1.168\\\hline
\multirow{2}{1.4cm}{cup} & RoBERTa & 0.354 & 2.942 & 1.483\\
& Random &  0.357 & 2.724 & 1.547\\
\bottomrule 
\end{tabular}\\
(b)
\caption{ (a) ZeroGaze results for RoBERTa and  fixed random embeddings. (b) Comparison of ZeroGaze results within categories.}
\label{table:zgrandom_results}
\end{table}
\def\subFigSz{0.32\linewidth}
\begin{figure}[ht!]
\centering
\includegraphics[width=\subFigSz]{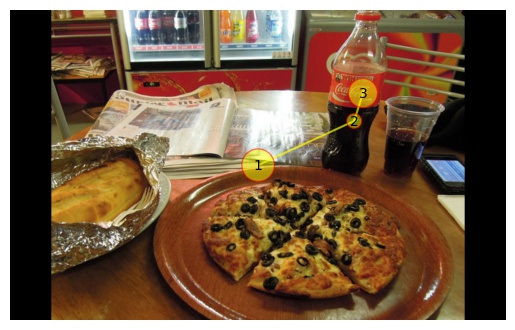} 
   \includegraphics[width=\subFigSz]{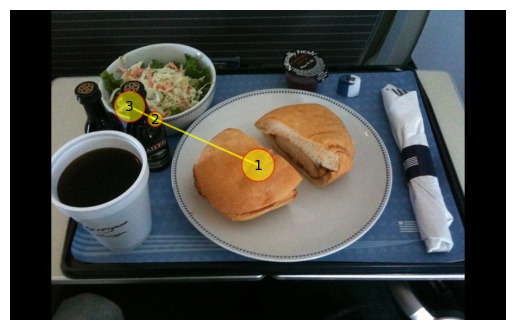}
  \includegraphics[width=\subFigSz]{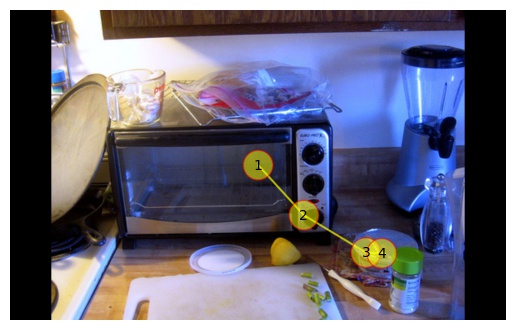}
\caption{Gazeformer often confuses a target with semantically similar objects under the ZeroGaze setting, e.g., being distracted by a bottle or bowl when the new target category is a cup.}
\label{fig:distractor}
\end{figure}

Table \ref{table:zgrandom_results}(a) shows that RoBERTa embeddings improve 
Gazeformer's performance under the ZeroGaze setting. Interestingly, Table \ref{table:zgrandom_results}(b) shows that language embeddings for different targets produce different benefits, with ``stop sign'' benefiting the most and ``cup'' the least. We speculate that this is due to how target words used in natural language differ in their semantic consistency. For example, the word ``cup'' is used in various contexts, often interchangeably with multiple different synonyms, e.g., drink, bowl, container. Indeed, we find that the model is often confused with other semantically similar objects when searching for a target; this is especially true when the target is a cup (Fig.~\ref{fig:distractor}). Note that even with fixed random embeddings, Gazeformer achieves slightly better performance with respect to the FFM and IRL models. We posit that this is due to the ability of the Gazeformer transformer encoder block to explicitly capture scene context which enables scene exploration - a considerable part of the search process. Nevertheless, semantic cues from the language model embeddings boost the performance of our model even further.

\subsection{Pixel Regression vs Grid Classification}

In \Tref{table:cls_results}, we demonstrate the impact of regressing the fixation co-ordinate parameters as compared to the more standard method of classifying amongst patches. We replace the regression step of Gazeformer with a classifier MLP 
that learns a probability distribution over the $20\times32$ patches (henceforth, called  \textit{Gazeformer-noReg}).

\setlength{\tabcolsep}{8pt}
\begin{table}[ht!]
\centering
\begin{tabular}{lccc}
\toprule 
& SS$\bm{\uparrow}$ & FED$\bm{\downarrow}$& NSS$\bm{\uparrow}$\\  
\midrule 
Gazeformer & \textbf{0.504} & \textbf{2.072} & \textbf{8.375}\\
Gazeformer-noReg & 0.477 & 2.158 & 7.545\\
\bottomrule 
\end{tabular}
\caption{Gazeformer performance compared to a model variant where pixel regression is replaced by patch classification (\textit{Gazeformer-noReg}). Results are from GazeTrain setting.}
\label{table:cls_results}
\end{table}

Table \ref{table:cls_results} shows that directly regressing the fixation locations provides significantly better performance, as hypothesized. 
Also note that even when the model estimates a grid probability distribution (similar to previous methods), the performance is still superior to baselines. This highlights the strength of our core architecture.

\section{Conclusion}

We introduced \textit{ZeroGaze}, a novel attention task aimed at scalability whereby a model must predict the fixation scanpaths for search targets despite having no prior search fixations available for training. We proposed a new multimodal, transformer-based model called \textit{Gazeformer} coupled with a novel scanpath prediction method, which not only showed impressive ZeroGaze results but also scaled well to uncommon and unknown 
target categories. Our model also achieved new state-of-the-art 
scanpath prediction performance in traditional target-present and target-absent search, while having faster inference speeds.
Because Gazeformer can generate fixation locations and durations for an entire scanpath in negligible time, it enables anticipation of \textit{where} and \textit{when} a person’s attention will shift, thus making it ideal for time-critical HCI applications. 
We expect that Gazeformer's scalability, effectiveness, and speed will fuel the use of gaze prediction models in HCI applications and products. In future work, 
we look forward to extensions of Gazeformer to other visual tasks like free-viewing and VQA. 

\myheading{Acknowledgement}. {\small  This project was supported by US National Science Foundation Awards IIS-1763981, IIS-2123920, NSDF DUE-2055406, and the SUNY2020 Infrastructure Transportation Security Center, and a gift from Adobe.}

{\small
\bibliographystyle{ieee_fullname}
\bibliography{egbib}
}
\newpage

\section{Appendix}
\label{sec:appendix}
\thispagestyle{empty}
\setlength{\tabcolsep}{4pt}
\begin{table}[h]
\centering
\begin{tabular}{l|cc|cc}
\toprule 
& \multicolumn{2}{c|}{SemSS$\bm{\uparrow}$} & \multicolumn{2}{c}{SemFED $\bm{\downarrow}$} \\  
 & w/o Dur & w/ Dur & w/o Dur & w/ Dur
 \\
 \Xhline{0.75pt}
IRL~\cite{yang2020predicting}& 0.285 & - & 4.558 & - \\
Chen \etal~\cite{chen2021predicting}  & 0.178 & 0.028 &  5.845 & 212.725 \\
FFM~\cite{yang2022target} & 0.282 & - & 3.132 & - \\ \hline
Gazeformer-noDur & \textbf{0.352} & -  & 2.630 & - \\
Gazeformer & \textbf{0.352} & \textbf{0.312} &  \textbf{2.586} & \textbf{10.932} \\
\bottomrule 
\end{tabular}\\
(a)\\

\begin{tabular}{l|cc|cc}
\toprule 
& \multicolumn{2}{c|}{SemSS$\bm{\uparrow}$} & \multicolumn{2}{c}{SemFED $\bm{\downarrow}$} \\  
 & w/o Dur & w/ Dur & w/o Dur & w/ Dur
 \\
 \Xhline{0.75pt}
Human & 0.522&0.433 & 1.720&8.389\\ \hline
IRL~\cite{yang2020predicting}  & 0.481&- & 2.259&- \\
Chen \etal~\cite{chen2021predicting} & 0.470&0.418 & 1.898&9.189\\
FFM~\cite{yang2022target}& 0.407&- & 2.425&- \\ \hline
Gazeformer-noDur & \textbf{0.496}&- & \textbf{1.861}&- \\
Gazeformer & 0.490&\underline{\textbf{0.456}} & 1.928 &\underline{\textbf{8.064}}\\
\bottomrule 
\end{tabular}\\
(b)
\caption{Semantic Sequence Score (SemSS) and Semantic Fixation Edit Distance (SemFED) comparison under (a) ZeroGaze setting, (b) traditional GazeTrain setting.  Best performance is highlighted in bold. Performances that exceed human consistency are underlined.}
\label{table:all_results_updated}
\end{table}

\setlength{\tabcolsep}{5pt}
\begin{table}[ht!]
\centering
\begin{tabular}{l|cc|cc}
\toprule 
& \multicolumn{2}{c|}{SemSS$\bm{\uparrow}$} & \multicolumn{2}{c}{SemFED $\bm{\downarrow}$} \\  
 & w/o Dur & w/ Dur & w/o Dur & w/ Dur
 \\
 \Xhline{0.75pt}
Human & 0.393 & 0.363 & 3.772 & 15.358 \\ \hline
IRL~\cite{yang2020predicting} & 0.310 & - & 5.171 & - \\
Chen \etal~\cite{chen2021predicting} & 0.356 & 0.341 & \underline{\textbf{3.349}} & \underline{13.600} \\
FFM~\cite{yang2022target} & 0.376 & - & \underline{3.414} & -  \\ \hline
Gazeformer-noDur &\textbf{0.381} & - & \underline{3.459} & - \\
Gazeformer  & 0.379 & \textbf{0.360} & \underline{3.390} & \underline{\textbf{13.449}}\\
\bottomrule 
\end{tabular}\\
(a)\\

\begin{tabular}{l|cc|cc}
\toprule 
& \multicolumn{2}{c|}{SemSS$\bm{\uparrow}$} & \multicolumn{2}{c}{SemFED $\bm{\downarrow}$} \\  
 & w/o Dur & w/ Dur & w/o Dur & w/ Dur
 \\
 \Xhline{0.75pt}
Human & 0.393 & 0.363 & 3.772 & 15.358 \\ \hline
IRL~\cite{yang2020predicting} & 0.339 & - & 5.031 & - \\
Chen \etal~\cite{chen2021predicting} & 0.315 & 0.303  & \underline{3.481} & \underline{\textbf{13.919}}\\
FFM~\cite{yang2022target} & 0.372 & - & 3.819 & - \\ \hline
Gazeformer-noDur & 0.381 & - & \underline{\textbf{3.430}} & - \\
Gazeformer & \underline{\textbf{0.394}} & \underline{\textbf{0.376}} & \underline{3.550} & \underline{13.960}\\
\bottomrule 
\end{tabular}\\
(b)
\caption{Semantic Sequence Score (SemSS) and Semantic Fixation Edit Distance (SemFED) comparison for models (a) trained with target-present data and tested on target-absent data, and (b) trained with target-absent data and tested on target-absent data. The best performance for each metric is highlighted in bold. Performance that exceeds human consistency is underlined.}
\label{table:ta_results_updated}
\end{table}
\setlength{\tabcolsep}{4pt}
\begin{table}[h]
\centering
\begin{tabular}{l|cc|cc}
\toprule 
& \multicolumn{2}{c|}{SemSS$\bm{\uparrow}$} & \multicolumn{2}{c}{SemFED $\bm{\downarrow}$} \\  
 & w/o Dur & w/ Dur & w/o Dur & w/ Dur
 \\
 \Xhline{0.75pt}
 IRL~\cite{yang2020predicting} & 0.310 & - & 5.220 & - \\
Chen \etal~\cite{chen2021predicting} & 0.077 & 0.016 & 5.769 & 150.626\\
FFM~\cite{yang2022target} & 0.242 & - & 4.834 & - \\ \hline
Gazeformer-noDur & \textbf{0.371} & - & 3.867 & - \\
Gazeformer & 0.363 & \textbf{0.350} & \textbf{3.812} & \textbf{14.837} \\
 \bottomrule 
\end{tabular}\\
\caption{Semantic Sequence Score (SemSS) and Semantic Fixation Edit Distance (SemFED) comparison for models trained with target-absent data and tested on target-absent data under the ZeroGaze setting. The best performance for each metric is highlighted in bold.}
\label{table:zs_ta_results_updated}
\end{table}
 Our previous implementation of Semantic Sequence Score (SemSS) and Semantic Fixation Edit Distance (SemFED) did not account for the padding that was applied to COCO images \textit{if}, after scaling, they did not fit the fixed resolution required for the collection of COCO-Search18 dataset~\cite{chen2021coco, yang2020predicting}. Hence, we update our implementation to accommodate this and report the updated SemSS and SemFED scores in Table~\ref{table:all_results_updated} (corresponding to Table~\ref{table:all_results} in main paper), Table~\ref{table:ta_results_updated} (corresponding to Table~\ref{table:ta_results} in main paper and Tables 3-4 in the supplemental), and Table~\ref{table:zs_ta_results_updated} (corresponding to Table 2 in the supplemental). We see the same trend in the updated scores as observed in our previously reported SemSS and SemFED scores. Gazeformer achieves state-of-the-art performance on the updated SemSS and SemFED metrics.
\end{document}